\documentclass[letterpaper]{article} 
\usepackage[submission]{aaai23}  
\usepackage{times}  
\usepackage{helvet}  
\usepackage{courier}  
\usepackage[hyphens]{url}  
\usepackage{graphicx} 
\usepackage{natbib}  
\usepackage{caption} 
\usepackage{algorithm}
\usepackage{algorithmic}

\usepackage{multirow}
\usepackage{subfigure}
\usepackage{amsmath,amssymb}
\usepackage{booktabs}
\usepackage[capitalize]{cleveref}
\crefname{section}{Sec.}{Secs.}
\Crefname{section}{Section}{Sections}
\Crefname{table}{Table}{Tables}
\crefname{table}{Tab.}{Tabs.}

\usepackage[table]{xcolor}

\DeclareMathOperator*{\argmin}{arg\,min}
\newcommand{\R}{\mathbb{R}}

\usepackage{newfloat}
\usepackage{listings}
\DeclareCaptionStyle{ruled}{labelfont=normalfont,labelsep=colon,strut=off} 
\floatstyle{ruled}
\newfloat{listing}{tb}{lst}{}
\floatname{listing}{Listing}
\title{A Simple Baseline for Multi-Camera 3D Object Detection}

\author{
    Yunpeng Zhang\textsuperscript{\rm 1},
    Wenzhao Zheng\textsuperscript{\rm 2},
    Zheng Zhu\textsuperscript{\rm 1},\\
    Guan Huang\textsuperscript{\rm 1},
    Jie Zhou\textsuperscript{\rm 2},
    Jiwen Lu\textsuperscript{\rm 2}
}
\affiliations{
    \textsuperscript{\rm 1} Phigent Robotics, \textsuperscript{\rm 2} Tsinghua University
}

\begin{document}

\maketitle

\begin{abstract}

3D object detection with surrounding cameras has been a promising direction for autonomous driving. In this paper, we present \textbf{SimMOD}, a \textbf{Sim}ple baseline for \textbf{M}ulti-camera \textbf{O}bject \textbf{D}etection, to solve the problem. To incorporate multi-view information as well as build upon previous efforts on monocular 3D object detection, the framework is built on sample-wise object proposals and designed to work in a two-stage manner. First, we extract multi-scale features and generate the perspective object proposals on each monocular image. Second, the multi-view proposals are aggregated and then iteratively refined with multi-view and multi-scale visual features in the DETR3D-style. The refined proposals are end-to-end decoded into the detection results. 
To further boost the performance, we incorporate the auxiliary branches alongside the proposal generation to enhance the feature learning. Also, we design the methods of target filtering and teacher forcing to promote the consistency of two-stage training.
We conduct extensive experiments on the 3D object detection benchmark of nuScenes to demonstrate the effectiveness of SimMOD and achieve new state-of-the-art performance. Code will be available at https://github.com/zhangyp15/SimMOD.

\end{abstract}

\section{Introduction}
The safety of self-driving vehicles strongly depends on an accurate and comprehensive understanding of the surroundings. As the core task of 3D scene understanding, 3D object detection aims to produce a 3D bounding box for each concerned object around the ego vehicle. Existing methods for 3D object detection usually rely on LiDAR sensors~\cite{pointpillars, pointrcnn} or stereo cameras~\cite{TL-stereo, disp-rcnn} to provide explicit distance information, but the high cost of LiDAR and the instability of stereo systems have hindered the practical application of autonomous driving.
Therefore, monocular 3D object detection has been considered a promising direction and received extensive attention in recent years.

While most existing methods~\cite{mono3d, fcos3d} focus on generating 3D bounding boxes based on one single monocular image, it is undoubtedly beneficial to incorporate the information from multiple surrounding cameras~\cite{detr3d, bevdet}, which are commonly available on modern self-driving vehicles as well as provided in recently released driving datasets~\cite{nuscenes, waymo}.
To fully exploit the multi-view images, DETR3D~\cite{detr3d} employs a set of learned queries to adaptively integrate multi-scale features by geometric projection. After the iterative processing, these queries are decoded into a set of 3D bounding boxes. However, DETR3D suffers from slow convergence and requires specialized pretraining to achieve promising performance. 

\begin{figure*}[t]
\centering
\includegraphics[width=\linewidth]{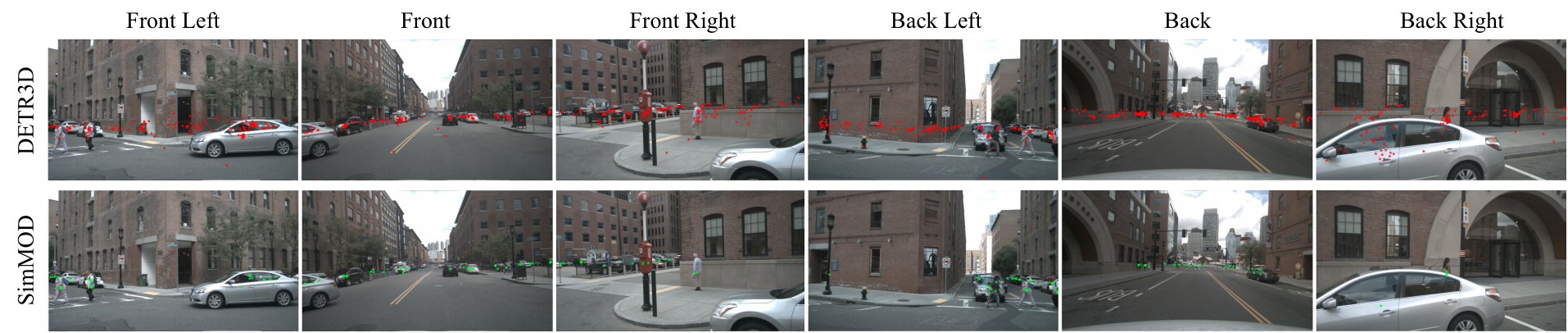}
\caption{\textbf{Qualitative comparison between fixed queries and sample-wise proposals}. As shown in the first row, the learned queries from DETR3D are scattered around the image and most lie in the background regions. In contrast, our sample-wise proposals (the second row) can effectively attend to the interested objects. (Best viewed in color.)
}
\label{fig:proposal_vs_query}
\end{figure*}

In this paper, we present SimMOD, a simple baseline for multi-camera 3D object detection, to serve as an effective and highly-extensible method to facilitate the development of multi-camera detection. SimMOD is formulated as a two-stage propose-and-fuse framework, which is compatible with existing methods for monocular 3D object detection and can also integrate information from multiple cameras. 
Specifically, we first utilize the image encoder to construct multi-scale feature maps for the multi-view images. The proposal head is then applied to distinguish the foreground objects and generate the sample-wise proposals with visual features, initial 3D positions, and view-level-aware encodings. 
In addition to the essential branches like the foreground classification and position estimation, we further incorporate a series of auxiliary branches to predict various visual attributes during the training stage. These auxiliary branches can significantly improve the learning of discriminative features. 
After the proposals are generated in every perspective view, we transform all proposals into the uniform ego-car coordinates and employ an iterative detection head to refine these proposals. This process includes the inter-proposal attention mechanism for interactions and the geometry-based feature sampling for aggregating multi-view information. 
Finally, the refined proposals are used to generate the detection results, which are supervised by the set-based losses to enable end-to-end predictions. 
Since the incomplete recall of generated proposals can hurt the bipartite matching, we also propose two techniques to provide consistent supervision. 
As shown in~\cref{fig:proposal_vs_query}, the sample-wise object proposals can focus on the interested objects, while most queries from DETR3D are scattered around the background regions. 

With extensive experiments, we verify the effectiveness of our proposed method on the nuScenes~\cite{nuscenes} dataset. Compared with the baseline DETR3D, SimMOD boosts the performance by 5.9\% NDS with ResNet-50 and 3.0\% NDS with ResNet-101. Also, SimMOD sets the new state-of-the-art performance on the nuScenes dataset.  

\section{Related Work}

\subsection{Image-based 2D Object Detection}

Modern image-based methods for 2D object detection can include two-stage and one-stage detectors. 
As the representation of two-stage methods, Faster-RCNN~\cite{faster_rcnn} uses the region proposal network to generate proposals and refine them with the RCNN~\cite{rcnn} head. By contrast, one-stage methods, like YOLO~\cite{yolo}, SSD~\cite{ssd}, FCOS~\cite{fcos} and so on, generate dense predictions in one shot. 
To eliminate the requirement of heuristic post-processing, the recent DETR~\cite{detr} provides an end-to-end detection framework based on transformer~\cite{transformer} and set-based losses. 
Despite its graceful design, DETR suffers from slow convergence and limited performance. To this end, Deformable DETR~\cite{deformable_DETR} proposes the deformable attention to efficiently utilize multi-scale features, while TSP~\cite{tsp_det} combines FCOS and the transformer-based set prediction for two-stage detection. 
Motivated by the two-stage DETR variants for 2D object detection, we present our two-stage framework for multi-camera 3D object detection. In the first stage, we recognize foreground objects in each perspective view and generate object proposals. For the second-stage refinement, we lift these proposals to the uniform 3D space and incorporate multi-view information for refinement and prediction. 

\subsection{Monocular 3D Object Detection}

With one single image as input, monocular 3D object detection aims to predict the 3D bounding boxes and categories of the interested objects. A series of methods rely on the additional depth information to solve the problem, including pseudo-LiDAR~\cite{pseudo-lidar}, D4LCN~\cite{D4LCN}, and CaDDN~\cite{caddn}. Other methods directly predict the 3D locations~\cite{monodle, monoflex, monopair} or incorporate the geometric clues~\cite{deep3Dbox, PGD}. 
Recently, FCOS3D~\cite{fcos3d} extends the 2D detector FCOS~\cite{fcos} for monocular 3D object detection and reports the performance on the multi-camera nuScenes dataset~\cite{nuscenes}. When dealing with multi-camera systems, monocular methods separately process each image and merge the results of detection through heuristic post-processing. The formulation is inherently sub-optimal because it fails in utilizing multi-view information and cannot benefit from end-to-end training.

\subsection{Multi-camera 3D Object Detection}

Recent methods for multi-camera 3D object detection decompose into two streams. A line of methods~\cite{bevdet, BEVFormer, zhang2022beverse} focuses on learning the Bird-Eye-View representations from images and then detects objects in BEV. Though these methods can construct the dense BEV representations for wide extensions, they can suffer from the accumulated errors in the view transformation. 
The other stream of methods still works on the level of objects. DETR3D~\cite{detr3d} learns a set of object queries, which can extract multi-camera features through the geometric projections of their reference points. However, the object queries of DETR3D are simply parameterized embeddings and cannot flexibly adapt to the different input samples. 
In this paper, we take DETR3D as our baseline and further propose a two-stage framework where high-quality and sample-dependent proposals are first generated and then refined with multi-view features in the second stage. We demonstrate the proposed method greatly outperforms the baseline DETR3D at the convergence rate and detection performance.

\begin{figure*}[t]
\centering
\includegraphics[width=\linewidth]{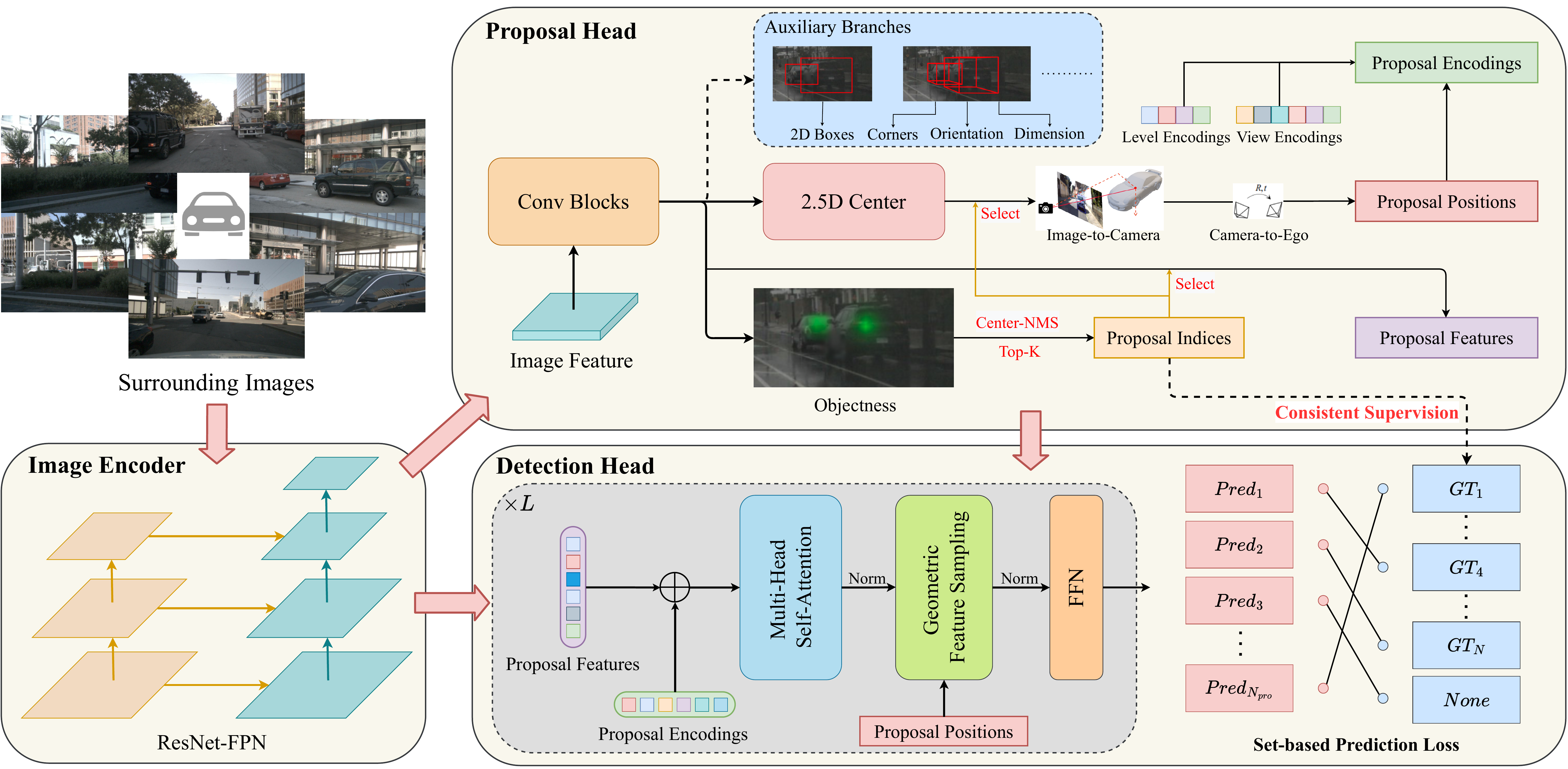}
\caption{\textbf{The overall framework of SimMOD}. With the surrounding images as input, SimMOD first extracts multi-scale feature maps with the image encoder, which consists of the backbone and the feature pyramid network. Next, the proposal head processes each feature map to generate the object proposals, including the features, positions, and encodings. Finally, the multi-view and multi-scale proposals are collected in the ego-car coordinates and iteratively refined. The set-based detection loss is applied for end-to-end predictions. (Best viewed in color.)}
\label{fig:framework}
\end{figure*}

\section{Approach}

\subsection{Overview}

Formally, the input of the overall framework includes a set of $M$ surrounding images $\mathcal{I} = \{\mathbf{I}_{1:M}\} \subset \R^{\mathrm{H} \times \mathrm{W}\times3}$ and their corresponding camera calibration matrices $\mathcal{K} = \{ \mathbf{K}_{1:M} \} \subset \R^{3\times3} $ and $\mathcal{T} = \{ \mathbf{T}_{1:M} \} \subset \R^{4\times4}$. The matrix $\mathbf{K}_j$ refers to the intrinsic matrix of camera $j$ and $\mathbf{T}_j$ refers to the transformation matrix from the coordinate in camera $j$ to the top-view LiDAR coordinate. As the learning targets, the ground-truth 3D bounding boxes are defined as $\mathcal{B} = \{\mathbf{b}_1, \ldots, \mathbf{b}_N\} \subset \R^{9}$ and their categorical labels are denoted as $\mathcal{C} = \{ c_1, \ldots, c_N \} \subset \mathcal{Z}$. Each box $\mathbf{b} = \left(x, y, z, w, l, h, \theta, v_x, v_y\right)$ contains the position, size, yaw, and velocity. The number of categories is denoted as $N_c$ and $N_c = 10$ for the nuScenes detection benchmark~\cite{nuscenes}. For convenience, we use $\mathcal{H}$ to transform the normal coordinate into the homogeneous coordinate and $\mathcal{H}^{-1}$ for the inverse. As illustrated in~\cref{fig:framework}, the proposed framework first extracts multi-scale features from the input multi-view images. Then the proposal head generates multi-view proposals and lifts them into the uniform 3D space. Finally, the detection head iteratively aggregates multi-view information and refines the proposals for the final prediction. The model is end-to-end optimized with two-stage and consistent supervision.

\subsection{Image Encoder}

For a fair comparison, we follow DETR3D~\cite{detr3d} to build the image-view encoder with the classical ResNet~\cite{resnet} and FPN~\cite{FPN}. The deformable convolutions~\cite{DCN} are also applied in the last two stages of the backbone for a better trade-off between accuracy and efficiency. Considering the variable scales of interested objects, the encoder can generate multi-scale feature maps for the generation and refinement of object proposals. Specifically, we use four levels with strides (8, 16, 32, 64) for the experiments. 

\subsection{Proposal Head}

The proposal head consists of four shared convolutional blocks and multiple small heads for different targets. Note that the proposal head is shared across different levels of feature maps. As for the distribution of objects to different feature levels and points, we follow the same settings as FCOS3D~\cite{fcos3d}. The proposal head is intended to distinguish foreground objects in the image domain, produce initial guesses about the 3D positions of these objects, and provide effective features for further refinement. We first introduce the generation of proposals and then elaborate on the design of proposal encodings. Finally, the auxiliary branches are incorporated to improve feature learning.

\paragraph{Proposal generation.}

The branches for generating proposals include classification, centerness, offset, and depth. The first two branches are combined to determine whether a point on the feature map corresponds to certain object centers, while the last two branches predict the 2.5D center, which can be lifted into the 3D position with camera matrices. Formally, we denote the predicted classification distribution as $ \mathbf{P}_{cls} \in \R^{N_c \times H' \times W'}$ and centerness as $ \mathbf{P}_{ctr} \in \R^{H' \times W'}$, where $(H', W')$ refers to sizes of the specific feature map. The objectness map $ \mathbf{P}_{obj} \in \R^{H' \times W'}$ can be computed as
\begin{equation}
\label{equ:object-ness}
\mathbf{P}_{obj} = \max \left( \mathbf{P}_{cls}, \textit{dim} = 0 \right) \odot \mathbf{P}_{ctr}
\end{equation}
where $\odot$ refers to the Hadamard product. Then we apply a $3 \times 3$ max-pooling on the objectness map to remove duplicates. Finally, the $N_{pro}$ top-scored positions are selected as object proposals. For each proposal, we assume it lies at camera $j$ and denote its corresponding pixel on the input image as $\left(u, v \right)$, the predicted offset as $\left(\Delta u, \Delta v\right)$, and the depth as $d$. The proposal position $(x, y, z)$ is computed as 
\begin{equation}
\label{equ:img_to_lidar}
(x, y, z, 1)^T = \mathbf{T}_j \mathcal{H} \left( {\mathbf{K}_j}^{-1} \left[ u + \Delta u, v + \Delta v, 1 \right]^T * d \right)
\end{equation}
where the image-view proposal is first lifted into the camera coordinate and then transformed to the top-view ego coordinate. The function $\mathcal{H}$ is used to get the homogeneous coordinate. In this way, the perspective proposals produced from each surrounding image are lifted into the same 3D coordinate system. We use $\mathbf{X}_{pro} \in \R^{N_{pro} \times 3}$ to denote the proposal positions. 
To further extract the proposal features for later refinement, we concatenate the corresponding features at the classification and regression branches and reduce the channels to get the proposal features $\mathbf{F}_{pro} \in \R^{N_{pro} \times C}$, where $C$ is the number of feature channels.

\paragraph{Proposal encodings.}

To further embed the positional information, we incorporate the corresponding camera-view and feature-level with the proposal positions to produce the proposal encodings for the subsequent refinement. Specifically, we define the learnable level encodings $\mathbf{E}_{level} \in \R^{N_{level} \times C} $ and view encodings $\mathbf{E}_{view} \in \R^{M \times C} $, where $N_{level}$ and $M$ refer to the numbers of feature levels and surrounding cameras. With the generated proposals, we select their corresponding level encodings and view encodings, concatenate them with the proposal positions, and linearly project to the proposal encodings $\mathbf{E}_{pro} \in \R^{N_{pro} \times 3}$. 

\paragraph{Auxiliary branches.}

The auxiliary branches aim to improve the object-awareness of preceding features by regressing other visual attributes of the objects. Since these branches only exist during the training stage, the computational burden for inference is not increased. We include the regression of the 2D bounding box, eight projected corners of the 3D bounding box, rotation, 3D size, and velocity as the auxiliary tasks. The supervision from these extra tasks not only improves the perception of fine-grained details like object boundaries and key-points, but also forces the network to learn high-level 3D information.

\subsection{Detection Head}

Given the multi-scale features and multi-view object proposals, the detection head motivated by DETR3D~\cite{detr3d} is designed to iteratively refine the proposals and generate predictions. 
Analogously to \cite{detr, detr3d}, the detection head includes $L$ iterative layers. Each layer contains the interaction among proposals and the interaction between proposals and image features. 
Formally, we use $\mathbf{F}_{I}^{jk}$ to represent the $k_{th}$ level feature map from the image of camera $j$. 
The proposal features and positions are denoted as $\mathbf{F}_{pro}$ and $\mathbf{X}_{pro}$. Three steps during each iteration are introduced in the following paragraphs. 

\paragraph{Interaction between proposals and images.} The interaction between proposals and images is important in two aspects. On the one hand, proposals generated from single-view images can further access the visual features from other views. On the other hand, the proposal features should be updated with the refinement of proposal positions during the iterations. To sample relevant features from multi-view and multi-scale feature maps, we project the proposal positions onto all these feature maps and use bi-linear interpolation to derive corresponding features. We use $\mathbf{f}_{I}^{jk}$ to represent the sampled features at $\mathbf{F}_{I}^{jk}$, and $\sigma^{jk}$ for the binary indicator to reflect whether the projected pixel is valid. Then the aggregated feature is given by:
\begin{equation}
    \label{equ:combine_proj_feat}
    \mathbf{f}_{pro}' = \mathbf{f}_{pro} + \frac{1}{\sum_{j} \sum_{k} \sigma^{jk}} \sum_{j} \sum_{k} \sigma^{jk} \mathbf{f}_{I}^{jk}
\end{equation}
Applying the aggregation in~\cref{equ:combine_proj_feat} for every proposal can get the updated proposal features $\mathbf{F}_{pro}'$.

\paragraph{Interaction among proposals.} The interaction among proposals is essential for removing duplicates and enabling end-to-end object detection. The multi-head attention mechanism~\cite{transformer} is utilized to realize the interaction. Specifically, we first compute the proposal encodings $\mathbf{E}_{pro} \in \R^{N_{pro} \times C}$ to incorporate the information of positions, feature levels, and camera views. Then we add these encodings to the proposal features and linearly project them to the queries $\mathbf{Q}_{pro}$, keys $\mathbf{K}_{pro}$, and values $\mathbf{V}_{pro}$. Finally, the attentive interaction is formulated as in~\cref{equ:self-attention}:
\begin{equation}
    \mathbf{F}''_{pro} = \mathbf{F}'_{pro} + \text{softmax} \left(\frac{\mathbf{Q}_{pro} \mathbf{K}_{pro}^T}{\sqrt{C}} \right) \mathbf{V}_{pro}
    \label{equ:self-attention}
\end{equation}

\paragraph{Decoding and updating proposals.} At the end of each iteration, we predict the 3D bounding box $\hat{\mathbf{b}}_i$ and its categorical distribution $\hat{\mathbf{c}}_i$ with two linear layers. The regression target of $\hat{\mathbf{b}}_i$ is defined as $\left(\Delta x, \Delta y, \Delta z, w, l, h, \sin \theta, \cos \theta, v_x, v_y \right)$, where $\left(\Delta x, \Delta y, \Delta z\right)$ is the residual vector between current proposal positions and their matched ground-truth boxes. With proposal features updated with the above interactions, we also refine the proposal positions by adding the predicted residual vectors.

\begin{table*}[t] 
\centering
\caption{Comparison to state-of-the-arts on the nuScenes validation set. \dag: initialized from a FCOS3D checkpoint.}
\label{tab:val}
\resizebox{\linewidth}{!}{
\begin{tabular}{lcccccccc}
\toprule
Method & Backbone & NDS $\uparrow$ & mAP $\uparrow$ & mATE $\downarrow$ & mASE $\downarrow$ & mAOE $\downarrow$ & mAVE $\downarrow$ & mAAE $\downarrow$ \\
\midrule

FCOS3D
~\cite{fcos3d} 
& \multirow{7}{*}{R101}
& 0.373 & 0.299 & 0.785 & 0.268 & 0.557 & 1.396 & \textbf{0.154} \\

DETR3D 
~\cite{detr3d} 
& & 0.374 & 0.303 & 0.860 & 0.278 & 0.437 & 0.967 & 0.235 \\

PGD
~\cite{PGD}
& & 0.409 & 0.336 & 0.732 & \textbf{0.263} & 0.423 & 1.285 & 0.172 \\

BEVDet
~\cite{bevdet} 
& & 0.396 & 0.330 & \textbf{0.702} & 0.272 & 0.534 & 0.932 & 0.251 \\

Ego3RT
~\cite{Ego3RT} 
& & 0.409 & 0.355 & 0.714 & 0.275 & 0.421 & 0.988 & 0.292 \\

PETR 
~\cite{PETR}
& & 0.421 & \textbf{0.357} & 0.710 & 0.270 & 0.490 & 0.885 & 0.224 \\

\rowcolor{green!25}
SimMOD 
& & \textbf{0.435} & 0.351 & 0.717 & 0.267 & \textbf{0.388} & \textbf{0.849} & 0.187 \\

\midrule

FCOS3D\dag 
~\cite{fcos3d}
& \multirow{8}{*}{R101} & {0.393} & {0.321} & {0.746} & {0.265} & {0.503} & {1.351} & \textbf{0.160} \\

PGD\dag
~\cite{PGD}
& & 0.425 & 0.358 & 0.667 & \textbf{0.264} &	0.435 & 1.278 &	0.177 \\

DETR3D\dag 
~\cite{detr3d}
& & {0.425} & {0.346} & {0.773} & {0.268} & {0.383} & {0.842} & {0.216} \\

Ego3RT\dag 
~\cite{Ego3RT} 
& & 0.450 & \textbf{0.375} & \textbf{0.657} & 0.268 & 0.391 & 0.850 & 0.206\\

PolarDETR\dag  
~\cite{PolarDETR}
& & 0.444 & 0.365 & 0.742 & 0.269 & 0.350 & 0.829 & 0.197 \\

BEVFormer-S\dag
~\cite{BEVFormer}
& & 0.448 & \textbf{0.375} & 0.725 & 0.272 & 0.391 & 0.802 & 0.200 \\

PETR\dag
~\cite{PETR}
& & 0.442 & 0.370 & 0.711 & 0.267 & 0.383 & 0.865 & 0.190 \\

\rowcolor{green!25}
SimMOD\dag & 
& \textbf{0.455} & 0.366 & 0.698 & \textbf{0.264} & \textbf{0.340} & \textbf{0.784} & 0.197\\

\bottomrule
\end{tabular}}
\end{table*}

\begin{table}[t]
\centering
\caption{Comparisons in the overlap regions. All methods use ResNet-101 as the backbone. \dag: initialized from a FCOS3D checkpoint.}
\begin{tabular}{lccccc}
\toprule
Method & NDS $\uparrow$ & mAP $\uparrow$ & mATE $\downarrow$ & mAOE $\downarrow$\\
\midrule 
FCOS3D & 0.317 & 0.213 & 0.841 & 0.604 \\
DETR3D & 0.356 & 0.231 & 0.825 & \textbf{0.400} \\
SimMOD & \textbf{0.394} & \textbf{0.274} & \textbf{0.789} & 0.437 \\
\midrule
FCOS3D\dag & 0.329 & 0.229 & 0.816 & 0.571 \\
DETR3D\dag & 0.384 & 0.268 & 0.807 & 0.453 \\
SimMOD\dag & \textbf{0.424} & \textbf{0.297} & \textbf{0.725} & \textbf{0.325}\\
\bottomrule
\end{tabular}
\label{tab:overlap}
\end{table}

\begin{table}[t]
\centering
\caption{The scalability of SimMOD with different image backbones. \dag: initialized from a FCOS3D checkpoint.}
\label{tab:backbones}
\resizebox{1.0\linewidth}{!}{
\begin{tabular}{cccccc}
\toprule
Method & Backbone & NDS & mAP & mATE & mAOE \\
\midrule 
CenterNet & DLA34 & 0.328 & \textbf{0.306} & \textbf{0.716} & 0.609 \\
SimMOD & DLA34 & \textbf{0.386} & 0.294 & 0.780 & \textbf{0.468} \\
\midrule
DETR3D & R50 & 0.373 & 0.302 & 0.811 & 0.493 \\
SimMOD & R50 & \textbf{0.432} & \textbf{0.339} & \textbf{0.727} & \textbf{0.356} \\
\midrule
FCOS3D & R101 & 0.373 & 0.299 & 0.785 & 0.557 \\
DETR3D & R101 & 0.374 & 0.303 & 0.860 & 0.437 \\
SimMOD & R101 & \textbf{0.435} & \textbf{0.351} & \textbf{0.717} & \textbf{0.388} \\
\midrule
DETR3D\dag & R101 & 0.425 & 0.346 & 0.773 & 0.383 \\
SimMOD\dag & R101 & \textbf{0.455} & \textbf{0.366} & \textbf{0.698} & \textbf{0.340} \\
\bottomrule
\end{tabular}}
\end{table}

\subsection{Consistent Supervision}
During the training stage, it is understandable that the generated proposals cannot achieve a complete recall of all ground-truth objects. Since the DETR-style set prediction loss is based on the bipartite matching between the refined proposals and the target objects, those objects missed by the proposals can hurt the matching and bring harmful gradients. To ensure the consistency between proposals and targets, we propose the following two techniques, including target filtering and teacher forcing.

\paragraph{Target filtering.} A natural way to maintain consistency is to remove the missed target objects for the bipartite matching. For each proposal, it can be matched to a target object if the proposal lies within the 2D bounding box. Then we compute the union of the matched objects from all proposals, which serves as the supervision for the second stage. 

\paragraph{Teacher forcing.} When training the recurrent neural networks, the method of teacher forcing randomly uses the ground-truth state, rather than the predicted state, as the next input, which can avoid the propagation of errors and stabilize the early training. To this end, we are motivated to randomly replace the predicted objectness map with the ground-truth so that all annotated objects can have their corresponding proposals. 

\begin{table*}[t]
\centering
\caption{Comparison to state-of-the-arts on the test set of nuScenes. L denotes LiDAR and C denotes camera. \ddag: the model is trained with external data.}
\label{tab:test}
\resizebox{\linewidth}{!}{
\begin{tabular}{lccccccccc}
\toprule
Method & Modality & Backbone & NDS $\uparrow$ & mAP $\uparrow$ & mATE $\downarrow$ & mASE $\downarrow$ & mAOE $\downarrow$ & mAVE $\downarrow$ & mAAE $\downarrow$ \\
\midrule 
PointPillars
& L & - & 0.453 & 0.305 & 0.517 & 0.290 & 0.500 & 0.316 & 0.368\\
CenterPoint
& L & - & 0.714 & 0.671 & 0.249 & 0.236 & 0.350 & 0.250 & 0.136\\
\midrule 
FCOS3D
& C & R101 & 0.428 & 0.358 & 0.690 & 0.249 & 0.452 & 1.434 & 0.124 \\
PGD
& C & R101 & 0.448 & 0.386 & 0.626 & 0.245 & 0.451 & 1.509 & 0.127 \\ 
Ego3RT 
& C & R101 & 0.443 & 0.389 & 0.599 & 0.268 & 0.470 & 1.169 & 0.172 \\
PETR 
& C & R101 & 0.455 & 0.391 & 0.647 & 0.251 & 0.433 & 0.933 & 0.143 \\
BEVFormer-S
& C & R101 & 0.462 & 0.409 & 0.650 & 0.261 & 0.439 & 0.925 & 0.147 \\
Graph-DETR3D
& C & R101 & 0.472 & 0.418 & 0.668 & 0.250 & 0.440 & 0.876 & 0.139 \\

\rowcolor{green!25}
SimMOD & C & R101 & 0.464 & 0.382 & 0.623 & 0.252 & 0.394 & 0.863 & 0.132\\

\midrule
DD3D\ddag
& C & V2-99 & 0.477 & 0.418 & 0.572 & 0.249 & 0.368 & 1.014 & 0.124 \\
DETR3D\ddag
& C & V2-99 & 0.479 & 0.412 & 0.641 & 0.255 & 0.394 & 0.845 & 0.133 \\
Ego3RT\ddag
& C & V2-99 & 0.473 & 0.425 & 0.549 & 0.264 & 0.433 & 1.014 & 0.145 \\
BEVDet\ddag
& C & V2-99 & 0.488 & 0.424 & 0.524 & 0.242 & 0.373 & 0.950 & 0.148 \\
PolarDETR\ddag
& C & V2-99 & 0.493 & 0.431 & 0.588 & 0.253 & 0.408 & 0.845 & 0.129 \\
Graph-DETR3D\ddag
& C & V2-99 & 0.495 & 0.425 & 0.621 & 0.251 & 0.386 & 0.790 & 0.128 \\
BEVFormer-S\ddag
& C & V2-99 & 0.495 & 0.435 & 0.589 & 0.254 & 0.402 & 0.842 & 0.131 \\
PETR\ddag
& C & V2-99 & 0.504 & 0.441 & 0.593 & 0.249 & 0.383 & 0.808 & 0.132 \\

\rowcolor{green!25}
SimMOD\ddag
& C & V2-99 & 0.494 & 0.417 & 0.570 & 0.248 & 0.387 & 0.813 & 0.126 \\
\bottomrule
\end{tabular}}
\end{table*}

\subsection{Loss Functions}

\paragraph{Loss for proposal head.} We follow the loss functions in FCOS3D~\cite{fcos3d} for training the proposal head. The classification branch is optimized with focal loss~\cite{focal_loss}, while the centerness branch is trained with binary cross-entropy loss. All other regression branches, including the offset, depth, 2D bounding box, size, and orientation, are supervised with smooth $L_1$ loss.
The integral loss for proposal generation is denoted as $\mathcal{L}_{pro}$.

\paragraph{Loss for detection head.} 

Following DETR3D, the discrepancy between the predicted and ground-truth objects is quantified as the set prediction loss for supervision. Without loss of generality, the ground-truth objects are denoted as $\mathcal{B} = \{ \mathbf{b}_i \}_{i=1}^N$ and their categorical labels as $\mathcal{C} = \{ c_i \}_{i=1}^N$. Also, the predicted bounding boxes and categorical distributions are represented with $\hat{\mathcal{B}} = \{ \hat{\mathbf{b}}_i \}_{i=1}^{N_{pro}}$ and  $ \hat{\mathcal{C}}  = \{ \hat{\mathbf{c}}_i \}_{i=1}^{N_{pro}}$. We pad the ground-truth boxes with $\varnothing$ up to $N_{pro}$ objects and use the Hungarian algorithm to compute the optimal assignment. Formally, finding the optimal bipartite matching requires searching for the optimal permutation $\hat{\sigma}$ of $N_{pro}$ elements with the lowest matching cost:
\begin{equation}
    \label{equ:optimal_perm}
    \hat{\sigma} = \argmin _{\sigma} \sum_{i=1}^{N_{pro}} \mathcal{L}_{\text{match}} \left( \mathbf{b}_i, c_i, \hat{\mathbf{b}}_{\sigma(i)}, \hat{\mathbf{c}}_{\sigma(i)} \right)
\end{equation}

We define the pair-wise matching cost $\mathcal{L}_{\text{match}}$ to consider both the classification and regression, as detailed in \cref{equ:match_cost}:
\begin{equation}
    \label{equ:match_cost}
    \mathcal{L}_{\text{match}} = -1_{\{ c_i \neq \varnothing \}} \hat{\mathbf{c}}_{\sigma(i)}(c_i) + 1_{\{ c_i \neq \varnothing \}} \lvert \mathbf{b}_i - \hat{\mathbf{b}}_{\sigma(i)} \rvert
\end{equation}

When the optimal assignment $\hat{\sigma}$ is computed, the set prediction loss is formulated as
\begin{equation}
    \label{equ:det_loss}
    \mathcal{L}_{\text{det}} = \sum_{i=1}^{N_{pro}} -\log\hat{\mathbf{c}}_{\hat{\sigma}(i)}(c_i) + 1_{\{c_i \neq \varnothing\}} \lvert \mathbf{b}_i - \hat{\mathbf{b}}_{\hat{\sigma}(i)} \rvert
\end{equation}

\paragraph{Total loss.}
The whole framework is end-to-end optimized with the two-stage loss $\mathcal{L}= \lambda \mathcal{L}_{pro} + \mathcal{L}_{\text{det}}$, where $\lambda = 1.0$ for our experiments. During the training stage, we follow existing methods~\cite{detr, detr3d} to deeply supervise the intermediate predictions from every layer of iteration. For inference, we only use the predictions from the last layer. 

\section{Experiments}

\subsection{Implementation Details}

\paragraph{Dataset.}

We conduct extensive experiments on the large-scale autonomous driving dataset nuScenes~\cite{nuscenes} to evaluate the proposed method. The nuScenes dataset provides 1000 sequences of different scenes collected in Boston and Singapore. Each sequence is about 20 seconds long and annotated with accurate 3D bounding boxes at 2Hz, contributing to a total of 1.4M object bounding boxes. These sequences are officially split into 700/150/150 ones for training, validation, and testing. For each sample, the images from six surrounding cameras and their camera calibrations are used as the input.

\paragraph{Evaluation metrics.}

Following the ofﬁcial evaluation protocol, the metrics include mean Average Precision (mAP) and a set of True Positive (TP) metrics, which contains the average translation error (ATE), average scale error (ASE), average orientation error (AOE), average velocity error (AVE), and average attribute error (AAE). 
Finally, the nuScenes detection score (NDS) is defined to consolidate the above metrics by computing a weighted sum as in \cref{equ:nds}:
\begin{equation}
    \text{NDS} = \frac{1}{10} \left[ 5\text{mAP} + \sum_{\text{mTP} \in \mathcal{TP}} \left(1 - \min{\left(1, \text{mTP}\right)} \right) \right]
    \label{equ:nds}
\end{equation}

\paragraph{Network architectures.} 
Unless specified, we use the ImageNet-pretrained ResNet-101~\cite{resnet} as the backbone network. We use the feature channel of 256 for multi-scale feature maps and proposal features. Within the proposal head, we define learnable and level-wise scale factors for regressing the offsets, depths, 2D boxes, and corners. We use the top-scored 600 proposals and filter low-scored proposals. The detection head contains six layers for iterative refinement.

\begin{figure}[t]
\includegraphics[width=\linewidth]{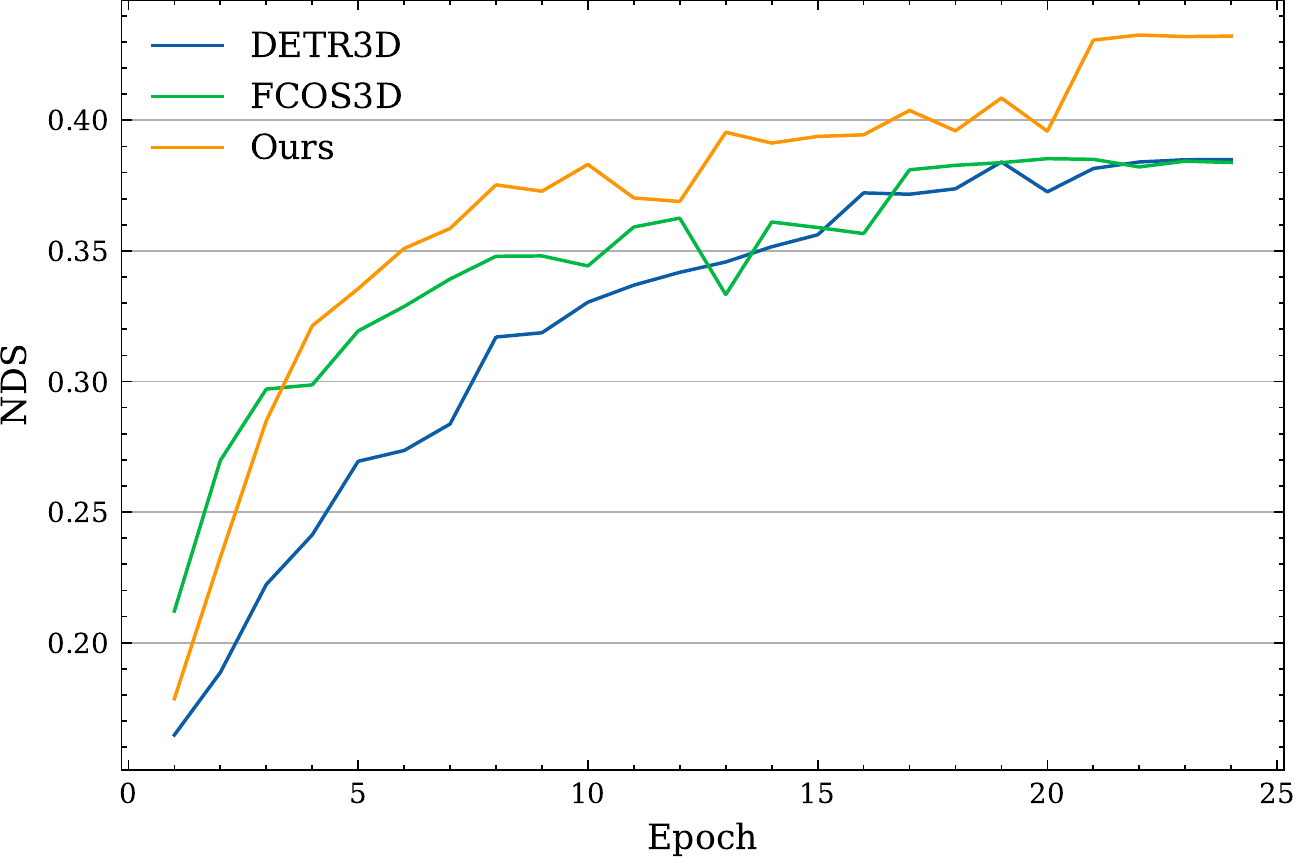}
\caption{The epoch-wise performance of FCOS3D, DETR3D, and our SimMOD on the nuScenes validation set.}
\label{fig:convergence}
\end{figure}

\paragraph{Training and inference.}

SimMOD is implemented based on mmdetection3d~\cite{mmdet3d2020}. The model is end-to-end trained with AdamW~\cite{adamw} optimizer for 24 epochs, with the initial learning rate as 2e-4 and weight decay as 0.01. Multi-step learning rate decay is applied. The input resolution is 1600 $\times$ 900. We train the model on 8 NVIDIA GeForce RTX 3090 GPUs with per-GPU batch size as 1. Following DETR3D~\cite{detr3d}, we use color distortion and GridMask~\cite{gridmask} for data augmentation.

\subsection{Benchmark Results}

\paragraph{NuScenes validation set.}
As shown in~\cref{tab:val}, we conduct a comprehensive comparison between the proposed SimMOD with existing state-of-the-art methods on the nuScenes validation set. 
When using the ImageNet-pretrained R101 as the backbone network, SimMOD with dynamic proposals greatly improves the baseline DETR3D by 6.1\% NDS and outperforms all existing methods. 
When initialized from the pretrained FCOS3D checkpoint, SimMOD achieves 45.5\% NDS and is still 3.0\% NDS higher than the baseline DETR3D. Despite the simplicity, the two-stage propose-and-fuse framework achieves the new state-of-the-art performance on the nuScenes validation set.

\paragraph{NuScenes test set.} 
For the nuScenes test set, we train SimMOD on the trainval split for 24 epochs with the strategy of CBGS~\cite{cbgs}. The single model is used for inference without test-time augmentations. Since our method only takes one single frame as input, we do not compare the methods with temporal information. 
As shown in~\cref{tab:test}, SimMOD with R101 achieves 46.4\% NDS and surpasses most state-of-the-art methods. When using the V2-99~\cite{centermask} backbone pretrained with depth estimation~\cite{dd3d}, SimMOD improves the baseline DETR3D~\cite{detr3d} by 1.5\% NDS and achieves comparable performance with the concurrent state-of-the-art methods, like PETR~\cite{PETR} and BEVFormer~\cite{BEVFormer}. Also, our SimMOD achieves better performance than the LiDAR-based method PointPillars~\cite{pointpillars}.

\begin{table}[t]
\centering
\caption{Ablation studies on sample-wise proposals.}
\label{tab:ablation_proposal}
\resizebox{\linewidth}{!}{
\begin{tabular}{lcccc}
\toprule
Method & NDS $\uparrow$ & mAP $\uparrow$ & mATE $\downarrow$ & mAOE $\downarrow$ \\
\midrule
Fixed & 0.362 & 0.293 & 0.855 & 0.502 \\
\midrule
Proposal & 0.401 & 0.347 & 0.742 & 0.479 \\
w. center-NMS & 0.409 & 0.349 & 0.726 & 0.487 \\
w. auxi (SimMOD) & \textbf{0.437} & \textbf{0.349} & \textbf{0.725} & \textbf{0.392} \\
\bottomrule
\end{tabular}}
\end{table}

\begin{table}[t]
\centering
\caption{Ablation studies on the consistent supervision.}
\label{tab:ablation_consistent}
\resizebox{\linewidth}{!}{
\begin{tabular}{cccccc}
\toprule
Filter & Teacher & NDS $\uparrow$ & mAP $\uparrow$ & mATE $\downarrow$ & mAOE $\downarrow$\\
\midrule 
& & 0.427 & 0.343 & 0.728 & 0.425 \\
\checkmark & & 0.436 & 0.347 & \textbf{0.724} & \textbf{0.383} \\
& \checkmark & 0.433 & 0.346 & 0.737 & 0.400 \\
\checkmark & \checkmark & \textbf{0.437} & \textbf{0.349} & 0.725 & 0.392 \\
\bottomrule
\end{tabular}}
\end{table}

\subsection{Performance in Overlap Regions}

With only limited information, it is difficult to localize the truncated objects at the border regions of images. As shown in~\cref{tab:overlap}, the early-fusion methods, including DETR3D and our SimMOD, can effectively utilize the multi-view information and outperform the monocular method FCOS3D. 
When initialized from the FCOS3D checkpoint, SimMOD with sample-wise proposals further improves the baseline DETR3D by 4.0\% NDS and 2.9\% mAP.

\subsection{Scalability with Different Backbones}

In~\cref{tab:backbones}, we investigate the scalability of SimMOD with different backbone networks, including DLA~\cite{dla} and various ResNets~\cite{resnet}. We can observe that SimMOD notably outperforms the baseline model under every backbone setting. Also, SimMOD with R50 achieves 43.2\% NDS and can outperform DETR3D with R101 (42.5\% NDS). 

\begin{table}[t]
\centering
\caption{Ablation studies on the proposal encoding.}
\label{tab:ablation_proposal_encoding}
\resizebox{\linewidth}{!}{
\begin{tabular}{cccccc}
\toprule
View & Level & NDS $\uparrow$ & mAP $\uparrow$ & mATE $\downarrow$ & mAOE $\downarrow$\\
\midrule 
& & 0.424 & 0.347 & 0.740 & \textbf{0.388}\\
\checkmark & & 0.434 & 0.347 & \textbf{0.718} & 0.401 \\
& \checkmark & 0.432 & \textbf{0.351} & 0.728 & 0.396 \\
\checkmark & \checkmark & \textbf{0.437} & 0.349 & 0.725 & 0.392 \\
\bottomrule
\end{tabular}}
\end{table}

\begin{table}[t]
\centering
\caption{Quantitative analysis of the two-stage weights.}
\label{tab:analysis_weights}
\resizebox{\linewidth}{!}{
\begin{tabular}{ccccc}
\toprule
Proposal Weight $\lambda$ & NDS $\uparrow$ & mAP $\uparrow$ & mATE $\downarrow$ & mAOE $\downarrow$\\
\midrule 
0.5 & 0.434 & 0.342 & 0.741 & 0.383 \\
1.0 & \textbf{0.437} & \textbf{0.349} & \textbf{0.725} & 0.392 \\
2.0 & 0.433 & 0.344 & 0.728 & \textbf{0.376} \\
\bottomrule
\end{tabular}}
\end{table}

\subsection{Training Convergence}

To compare the convergence rate and final performance of SimMOD and the baseline methods, we train three methods under the same schedule and summarize the results in~\cref{fig:convergence}. One can find that the monocular method FCOS3D~\cite{fcos} converges much faster than the top-down method DETR3D~\cite{detr3d}, while DETR3D~\cite{detr3d} has a higher performance limit at convergence. By contrast, our proposed two-stage paradigm can possess all their advantages.
On the one hand, SimMOD can benefit from the guidance in the image views and quickly learn to recognize foreground objects, contributing to faster convergence. On the other hand, the second stage can further utilize multi-scale and multi-view features to refine the proposals, pushing the performance limit to a higher level.

\begin{figure*}[t]
\centering
\includegraphics[width=1.0\linewidth]{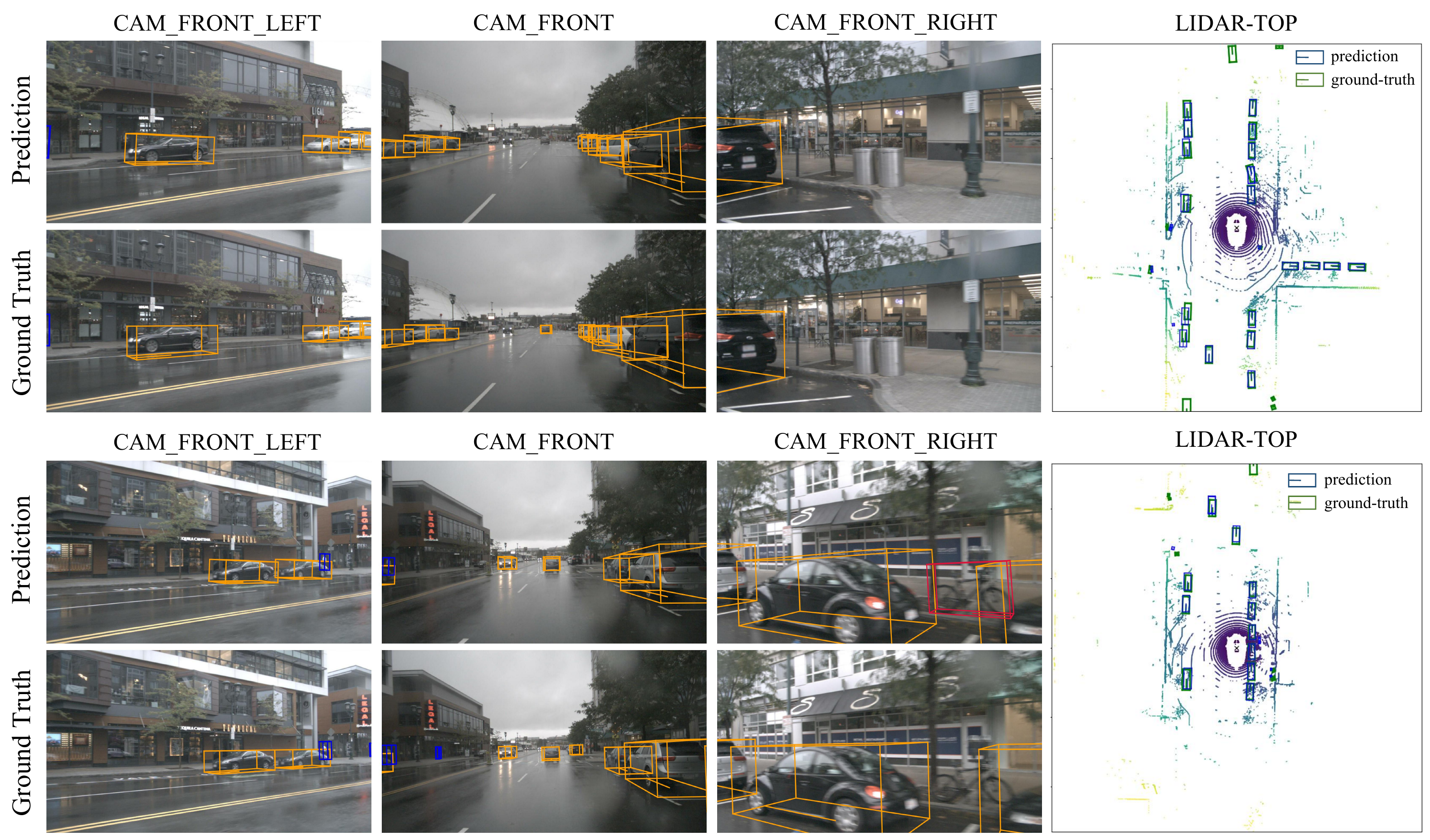}
\caption{\textbf{Qualitative visualization of the detection results}. 
We show the predicted and ground-truth boxes on the surrounding images (left) and the bird's eye view (right). For the bird's eye view, the predicted and ground-truth objects are shown in blue and green boxes. SimMOD can generate satisfactory detection results. (Best viewed in color)}
\label{fig:det_visualize}
\end{figure*}

\subsection{Ablation Studies}
\label{sec:ablation}

For ablation studies, we train SimMOD with ResNet-101 on 1/4 training data and initialize it from the pretrained FCOS3D checkpoint. 
We validate our designs in the following three aspects, including learning sample-wise proposals, promoting consistent supervision, and designing the proposal encodings. 

\paragraph{Sample-wise proposals.}

Since our core contribution is the two-stage propose-and-fuse framework, the design choices of proposal generation play a key role in improving the overall detection performance. As shown in~\cref{tab:ablation_proposal}, we first introduce the prediction of objectness map and depths to generate the proposals. Compared with the fixed queries, the minimal design with proposals can already improve NDS from 36.2\% to 40.1\%. 
Then, a simple 3$\times$3 max-pooling is conducted on the predicted objectness map to effectively remove duplicates and improve the performance by 0.8\% NDS. 
Finally, the auxiliary branches are applied to enhance the feature learning, which brings a significant boost of 2.8\% NDS. 

\begin{figure}[t]
\centering
\includegraphics[width=1.0\linewidth]{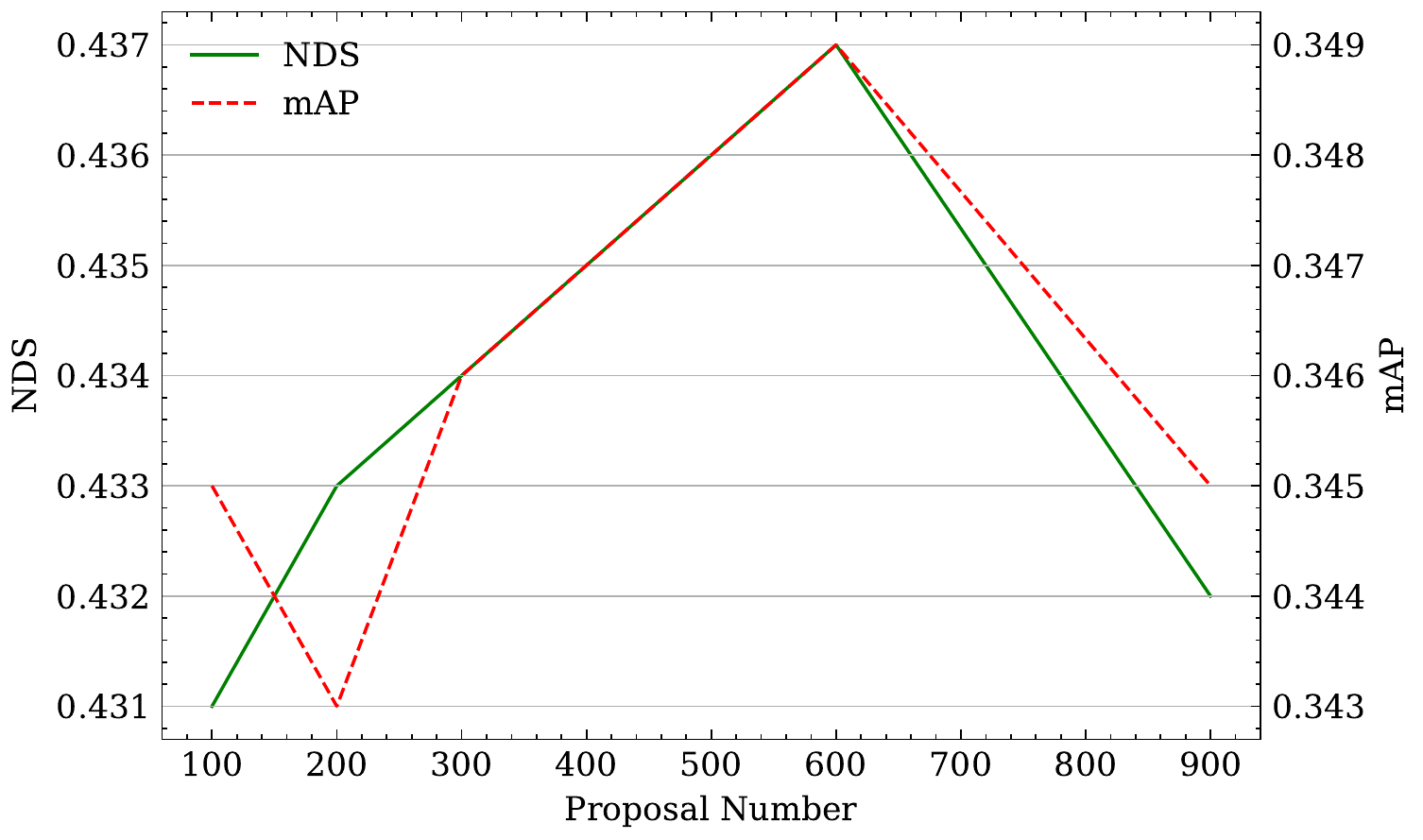}
\caption{Qualitative analysis on the number of proposals. Though the number of proposals varies widely, the performance of SimMOD is rather stable. }
\label{fig:tune_num_proposal}
\end{figure}

\paragraph{Consistent supervision.}

In~\cref{tab:ablation_consistent}, we verify the effectiveness of target filtering and teacher forcing in guaranteeing consistent supervision. We can observe that both techniques can mitigate inconsistency and improve performance. When jointly applied, the consistent supervision can outperform the default baseline by 1.0\% NDS. 

\paragraph{Proposal encoding.} 

Considering the proposal encoding, we validate the importance of incorporating the views and levels. As shown in~\cref{tab:ablation_proposal_encoding}, embedding the information of which camera and which level into the proposal can notably improve the overall performance. Since one object can potentially generate proposals at multiple views and levels, incorporating such information can help in removing duplicates and promoting inter-view interactions.

\subsection{Hyperparamter Sensitivity}

In this section, we investigate the influence of main hyperparameters, including the number of proposals and the weights for balancing the two-stage learning. Also, we use 1/4 of the training set for experiments.

\paragraph{Number of proposals.} In~\cref{fig:tune_num_proposal}, we show the performance of SimMOD with respect to different numbers of proposals. Thanks to the design of sample-wise proposals, SimMOD is relatively robust to the number of proposals. With only 100 proposals, SimMOD scores 43.1\% NDS, which is slightly worse than the 43.7\% NDS with 600 proposals. Besides, further increasing the number of proposals cannot improve the performance. 

\paragraph{Two-stage loss weights.} In~\cref{tab:analysis_weights}, we analyze the influence of tuning the weight $\lambda$ to balance the two-stage learning. Despite being a two-stage paradigm, SimMOD can achieve satisfactory performance with various values of $\lambda$.

\subsection{Qualitative Results}
As shown in \cref{fig:det_visualize}, we visualize the detection results in two challenging driving scenes. SimMOD can generate highly-accurate 3D bounding boxes for objects within a moderate range. However, it fails to detect some remote objects, which are relatively small in images.

\section{Conclusion}

In this paper, we present SimMOD: a simple two-stage propose-and-fuse framework for 3D object detection from multiple surrounding images. In the first stage, existing monocular detectors can be utilized to process single-view images and generate high-quality proposals. For the second stage, multi-view proposals are aggregated into the same 3D space and iteratively refined with attention mechanism and geometry-based feature sampling. 
Experimental results on the nuScenes detection benchmark demonstrate the superiority of SimMOD with new state-of-the-art performance.
We hope the proposed framework can serve as a simple and strong baseline for multi-camera 3D object detection.

\bibliography{arxiv.bib}

\end{document}